\documentclass{ecai}
\usepackage{times}
\usepackage{graphicx}
\usepackage{latexsym}

\usepackage{amsmath,amssymb,amsthm,mathtools,xspace}
\usepackage{enumerate}
\usepackage{xspace}
\usepackage{xcolor}
\usepackage{booktabs}

\usepackage{fvextra}
\usepackage{relsize}
\definecolor{coolblack}{rgb}{0.0, 0.18, 0.39}
\definecolor{darkblue}{rgb}{0.0, 0.0, 0.55}
\fvset{fontfamily=courier,fontsize=\relsize{-2},frame=lines,numbers=none,numbersep=4pt,commentchar=!,commandchars=\\\{\}}
\newcommand\atom{\Verb*[fontfamily=courier,fontsize=\relsize{-1}]}
\newcommand{\ActionSchema}[1]{\underline{\texttt{#1}}}

\newcommand{\tup}[1]{\langle #1 \rangle}

\newcommand{\mysmall}{}

\newcommand{\AMO}{\text{\mysmall At-Most-1\ }}
\newcommand{\ExactlyOne}{\text{\mysmall Exactly-1\ }}
\newcommand{\SLex}{\text{\mysmall Strict-Lex-Order\ }}

\newcommand{\Iff}{\Leftrightarrow}
\newcommand{\Implies}{\Rightarrow}

\newcommand{\MaxAtomArity}{\text{\mysmall max-arity}}

\newcommand{\ArityOf}[1]{\text{\mysmall arity(#1)}}

\newcommand{\metaf}{m}
\newcommand{\metao}{\nu}

\newcommand{\Using}{\text{\mysmall use}}
\newcommand{\PreZero}{\text{\mysmall p0}}
\newcommand{\PreOne}{\text{\mysmall p1}}
\newcommand{\EffZero}{\text{\mysmall e0}}
\newcommand{\EffOne}{\text{\mysmall e1}}
\newcommand{\Label}{\text{\mysmall label}}
\newcommand{\Arity}{\text{\mysmall arity}}
\newcommand{\Atom}{\text{\mysmall at}}
\newcommand{\Arg}{\text{\mysmall arg}}
 \newcommand{\Relevant}{\text{\mysmall argval}}
\newcommand{\StaticZero}{\text{\mysmall static0}}
\newcommand{\StaticOne}{\text{\mysmall static1}}

\newcommand{\Unary}{\text{\mysmall un}}
\newcommand{\Binary}{\text{\mysmall bin}}

\newcommand{\Head}[1]{\textcolor{darkblue}{#1}}

\newcommand{\Map}{\text{\mysmall mp}}
\newcommand{\Mapf}{\text{\mysmall mf}}
\newcommand{\Mapt}{\text{\mysmall mt}}
\newcommand{\Free}{\text{\mysmall free}}
\newcommand{\Appl}{\text{\mysmall appl}}

\newcommand{\Eq}{\text{\mysmall eq}}
\newcommand{\G}{\text{\mysmall g}}
\newcommand{\Ground}{\text{\mysmall gr}}
\newcommand{\Ord}{\text{\mysmall ord}}
\newcommand{\Gtuple}{\text{\mysmall gtuple}}
\newcommand{\ViolatedZero}{\text{\mysmall vio0}}
\newcommand{\ViolatedOne}{\text{\mysmall vio1}}
\newcommand{\PreZeroEq}{\text{\mysmall pre0eq}}
\newcommand{\PreOneEq}{\text{\mysmall pre1eq}}

\newcommand{\Omit}[1]{}

\newtheorem{thm}{Theorem}
\newtheorem{definition}[thm]{Definition}

\begin{document}
\allowdisplaybreaks

\title{Learning First-Order Symbolic  Representations for Planning from the Structure of the State Space}
\author{%
  Blai Bonet\institute{Universidad Sim\'on Bol\'{\i}var, Venezuela.  Email: bonet@usb.ve}
  \and
  Hector Geffner\institute{ICREA \& Universitat Pompeu Fabra, Spain. Email: hector.geffner@upf.edu}
}

\maketitle

\begin{abstract}
  One of the main obstacles for developing flexible AI systems is the split
  between data-based learners and model-based solvers. Solvers such as
  classical planners are very flexible and can deal with a variety of problem
  instances and goals but require first-order symbolic models. Data-based
  learners, on the other hand, are robust but do not produce such
  representations. In this work we address this split by showing how the
  \emph{first-order symbolic representations} that are used by planners can
  be learned \emph{from non-symbolic inputs} that encode the structure of
  the state space. The \emph{representation learning problem} is formulated as
  the problem of   inferring planning instances  over a common but unknown
  first-order domain that account for the structure of the observed state space.
  This means to infer a complete first-order representation (i.e.\ general action
  schemas, relational symbols, and objects) that explains the observed
  state space structures.
  The inference problem is cast as a two-level combinatorial search where
  the outer level searches for values of a small set of hyperparameters and
  the inner level, solved via  SAT, searches for a first-order symbolic model.
  The framework is shown to produce general and correct first-order representations
  for standard problems like Gripper, Blocksworld, and Hanoi from input graphs that 
  encode   the flat state-space structure of a single instance.
\end{abstract}

\section{INTRODUCTION}

Two of the main research threads  in AI  revolve around  the development of \emph{data-based learners}
capable of inferring  behavior and  functions from experience and data, and \emph{model-based solvers} capable of tackling well-defined
but intractable models like SAT, classical planning, and Bayesian networks.
Learners, and in particular deep learners, have  achieved considerable success but
result  in black boxes that do not have the flexibility, transparency, and generality
of their model-based counterparts \cite{josh,marcus1,pearl,darwiche,geffner:ijcai2018}.
Solvers, on  the other hand, require  models which are hard to build by hand.
This work  is aimed at bridging this gap by addressing  the problem of learning
first-order models from   data without using any  prior symbolic knowledge.

Almost all   existing approaches  for  learning representations for   acting and planning
fall into two camps  to be discussed below.  On the one hand,   methods that output
symbolic representations but which require symbolic representations in the input;
on the other,  methods that do not require symbolic inputs but which do not
produce them either. 
First-order representations structured in terms of  objects and relations  like PDDL
\cite{pddl,pddl:book,geffner:book}, however, have a number of benefits; in particular,
they are easier to understand,  and they   can  be easily  reused for defining
a variety of new instances and  goals. Representations like PDDL, however, are written by
hand; the challenge is to learn them  from  data.

In the proposed formulation, general first-order planning representations are learned from
graphs that encode the structure of the state space of one or more problem instances.
For this, the   \emph{representation learning problem} is  formulated as the problem of inferring
planning instances $P_i$ over a common, fully unknown, first-order domain $D$ (action schemas and predicate symbols)
such that the graphs $G(P_i)$ associated with the instances $P_i$ and the observed graphs $G_i$ are  \emph{structurally equivalent}.
Since the space of possible domains can be bounded by a number of hyperparameters with small values, such as the number of action schemas,
predicates, and arguments, the inference problem is cast as a two-level combinatorial search where the outer level looks for
the right value of the hyperparameters and the inner level, formulated and solved via SAT, looks for a first-order representation
that fits the hyperparameters and explains the input graphs.
Correct and general first-order models for domains like  Gripper, Blocksworld, and Hanoi are shown to be learned
from graphs that encode the flat  state-space structure of a single small instance.



\section{RELATED RESEARCH}

{Object-oriented MDPs} \cite{oo-mdp1}  and similar work in classical planning
\cite{yang:model-learning,review-learning-planning,sergio:model-recog},  build   {first-order model  representations} but starting with a first-order symbolic language,
or with information about the actions and their arguments \cite{locm}.
Inductive logic programming methods  \cite{ilp1} have been used for learning {general policies} but from symbolic encodings too
\cite{khardon:generalized,martin:generalized,fern:generalized}.
More recently, general policies have been learned using {deep learning} methods but also   starting with PDDL models \cite{trevizan:dl,sanner:dl,fern:dl,mausam:dl}.
The same holds for   methods for  learning   abstract planning  representations  \cite{bonet:aaai2019}. Other recent methods produce
PDDL models from  given  macro-actions (options) but these models are propositional and hence do not generalize  \cite{konidaris:jair}.

{Deep reinforcement learning} (DRL) methods \cite{atari}, on the other hand,
generate  policies  over  high-dimensional perceptual spaces like images,
{without using   any prior symbolic  knowledge} \cite{sid:sokoban,babyAI,pineau}.
Yet by not  constructing first-order   representations,  DRL methods lose the benefits  of transparency,
reusability, and compositionality \cite{marcus1,baroni:compositionality}.
Recent   work in {deep symbolic relational  reinforcement learning}  \cite{shanahan:review} attempts
to account for objects and relations through the use of attention mechanisms and loss functions  but
the {semantic and conceptual gap}  between the  {low-level techniques} and the {high-level  representations}
that are required remains  just too large.
Something similar occurs with work aimed at learning {low-dimensional
representations} that {disentangle  the factors of variations in the data}  \cite{bengio:entangled}.
The first-order  representations used in planning are low dimensional  but highly structured,
and it is  not clear that they can be learned in this way.
An alternative approach  produces first-order  representations using  a class of {variational autoencoders}
that provide  a low-dimensional encoding  of  the  images representing the states \cite{asai:latplan,asai:fol}.

\section{FORMULATION}

The proposed  formulation for learning planning representations from data departs from  existing approaches in
two  fundamental ways. First, unlike deep learning approaches, \emph{the representations are not learned from
images  associated with states but from the structure of the state space}. Second, unlike other methods
that deliver first-order symbolic planning representations, the \emph{proposed method does not assume
knowledge of the  action schemas, predicate symbols, or objects}; these are all learned from the input.
\emph{All the data required}  to learn planning representations in the four  domains  considered in the
experiments, Blocksworld, Towers of Hanoi, Gripper, and Grid, is shown in Fig.\,\ref{fig:graphs}.
In each case, the sole input is a labeled directed  graph that encodes  the structure of the state-space
associated with a small problem  instance, and the output is a  PDDL-like
representation made up of a general first-order domain with     action schemas and predicate symbols,
some of which are  possibly static, and  instance information describing  objects  and an initial situation.
No  goal information, however, is assumed in the input, and no goal information is produced in the output.
Further details about the inputs and outputs of the representation learning approach are described  below.

\subsection{Inputs: Labeled Graphs}

The inputs are  one or more labeled directed graphs that encode the structure of the state space of
one or several problem instances. The nodes of these graphs represent the  states and no information
about the contents or inner structure of them is provided or needed. Labels in the edges  represent action types;
e.g.,  action types in the  Gripper domain  distinguish three types of  actions:  moves, pick ups,
and drops. In the absence of labels,  all edges are assumed to have the same label. No other information
is provided as input or in the input graphs; in particular, no node is marked as an initial or goal state
(see Fig.~\ref{fig:graphs}). Initial states for each input graph are inferred indirectly, 
but nothing is inferred about goals.

In standard, tabular (model-free)  reinforcement learning (RL) schemes \cite{sutton:book}, the inputs are
traces made up of  states, actions, and rewards, and the task is to learn a policy for maximizing (discounted) expected reward,
not to learn factored symbolic representations of the actions and states. The same inputs are used
in model-based reinforcement learning methods where the policy is derived from a
flat model that is learned incrementally but  which does not transfer to other problems \cite{brafman:rmax}.

One way to understand the   labeled input graphs   used by  our method is as collections  of RL traces
organized as graphs  but without information about rewards and with the actions  replaced by less specific
action types or labels. Indeed, for a given node in an input graph, there may be zero, one, or many outgoing
edges labeled with the same action type. We assume however that input graphs  are \emph{complete}
in the sense that if they do not have an edge $(s,l,s')$ between two nodes $s$ and $s'$ with action label $l$,
it is because there is no trace that contains such a transition. Since the graphs required for building
the target representations are not large, a sufficient number of sampled traces can be used  to produce such
graphs. In the experiments, however, we built the input graphs by systematically expanding
the whole state space of a number  of small problem instances (in our case just one) from some
initial state. Formally, the input graphs  are tuples  $G\,{=}\,\tup{V,E,L}$, where the nodes $n$ in $V$
correspond to the different states, and the edges $(n,n')$ in $E$  with label $l \in L$, denoted $(n,l,n')$,
correspond to state transitions produced by an action with label $l$. For the sake of the presentation,
it is assumed that  all nodes in an  input graph can be reached from one or more nodes. The formulation, however,
does not require  this assumption.


\begin{figure*}[t]
  \centering
  \begin{tabular}{ccccccc}
    \resizebox{.40\columnwidth}{!}{\includegraphics{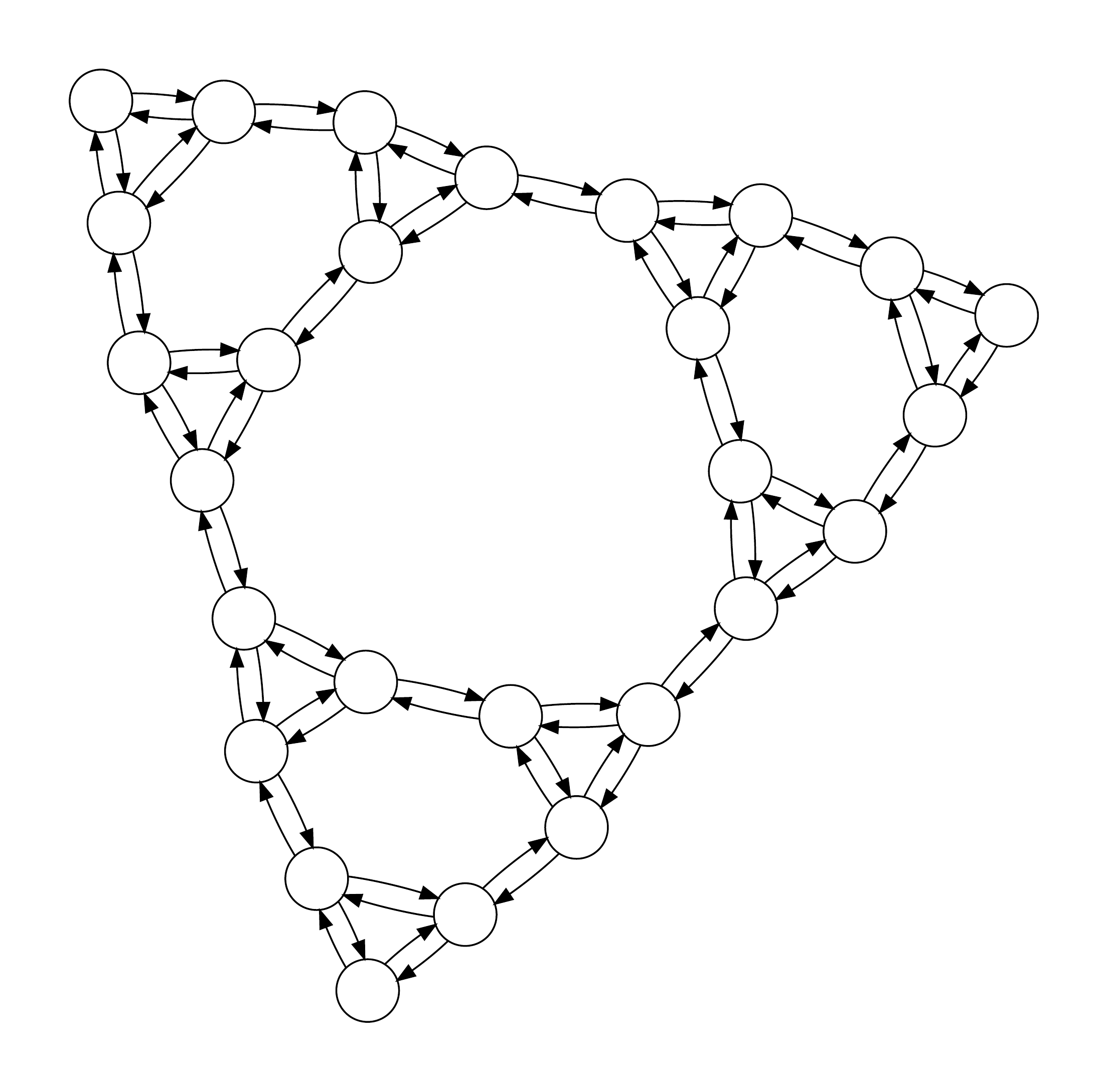}} &&
    \resizebox{.40\columnwidth}{!}{\includegraphics{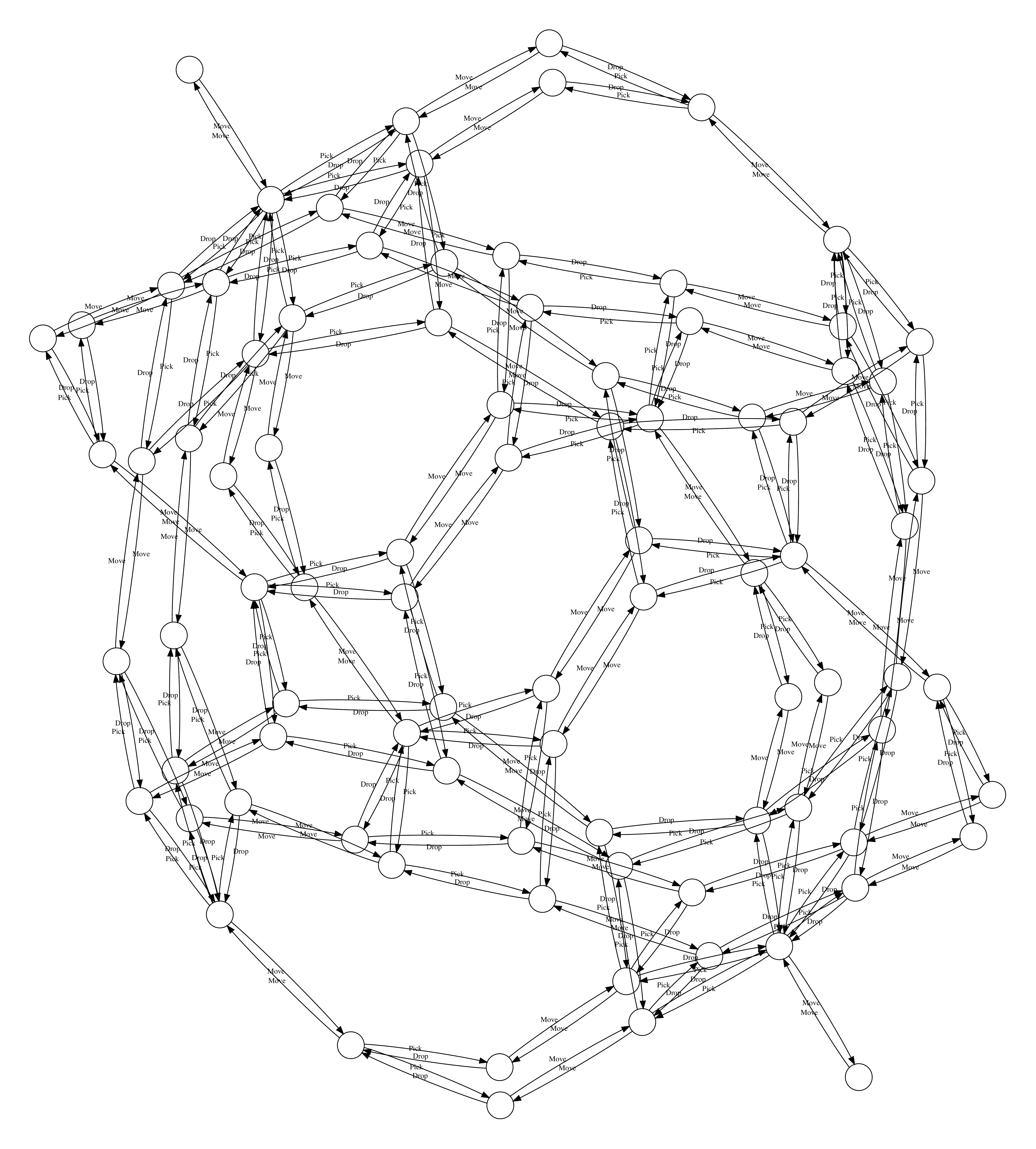}} &&
    \resizebox{!}{.43\columnwidth}{\includegraphics{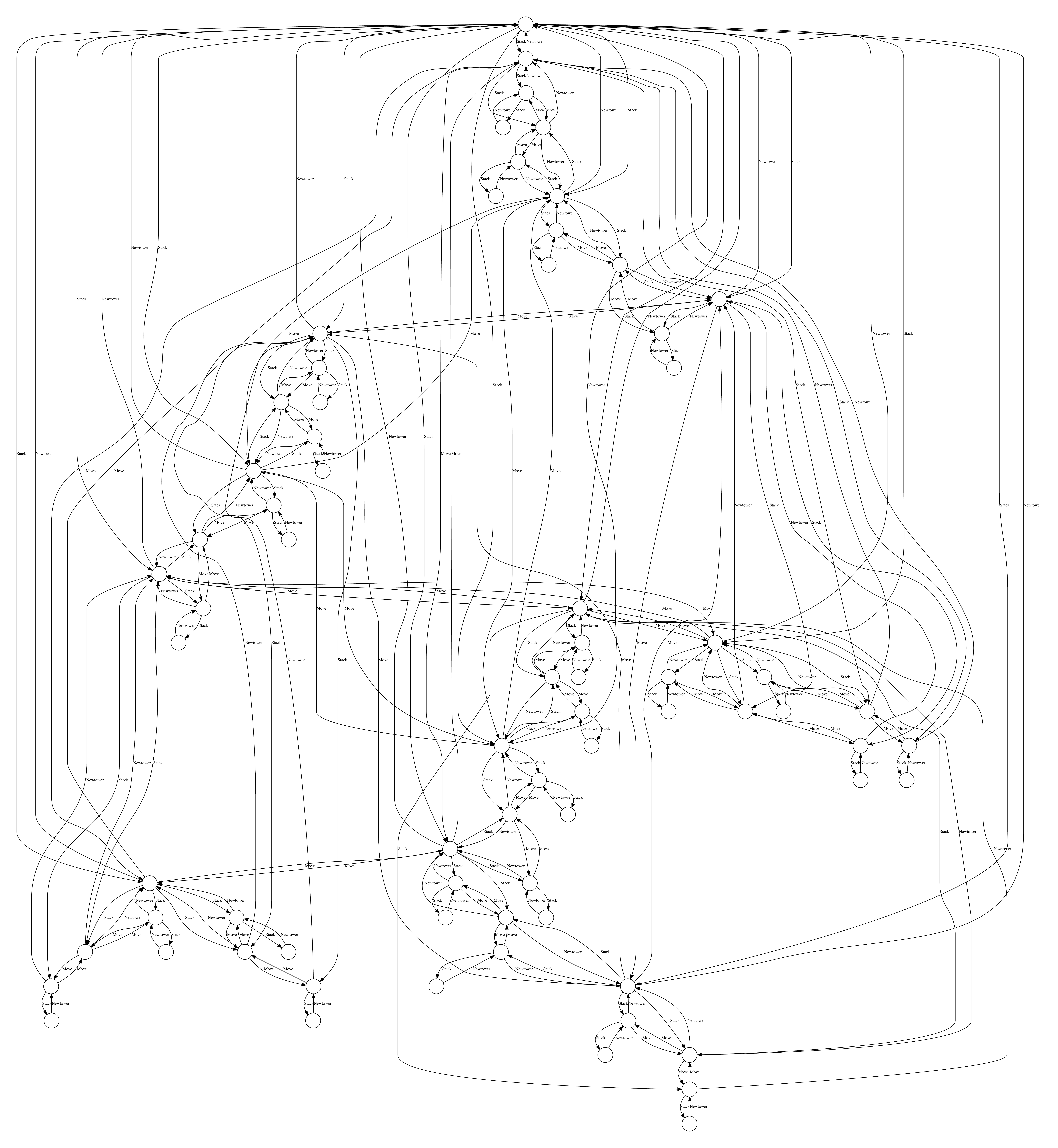}} &&
    \resizebox{!}{.43\columnwidth}{\includegraphics{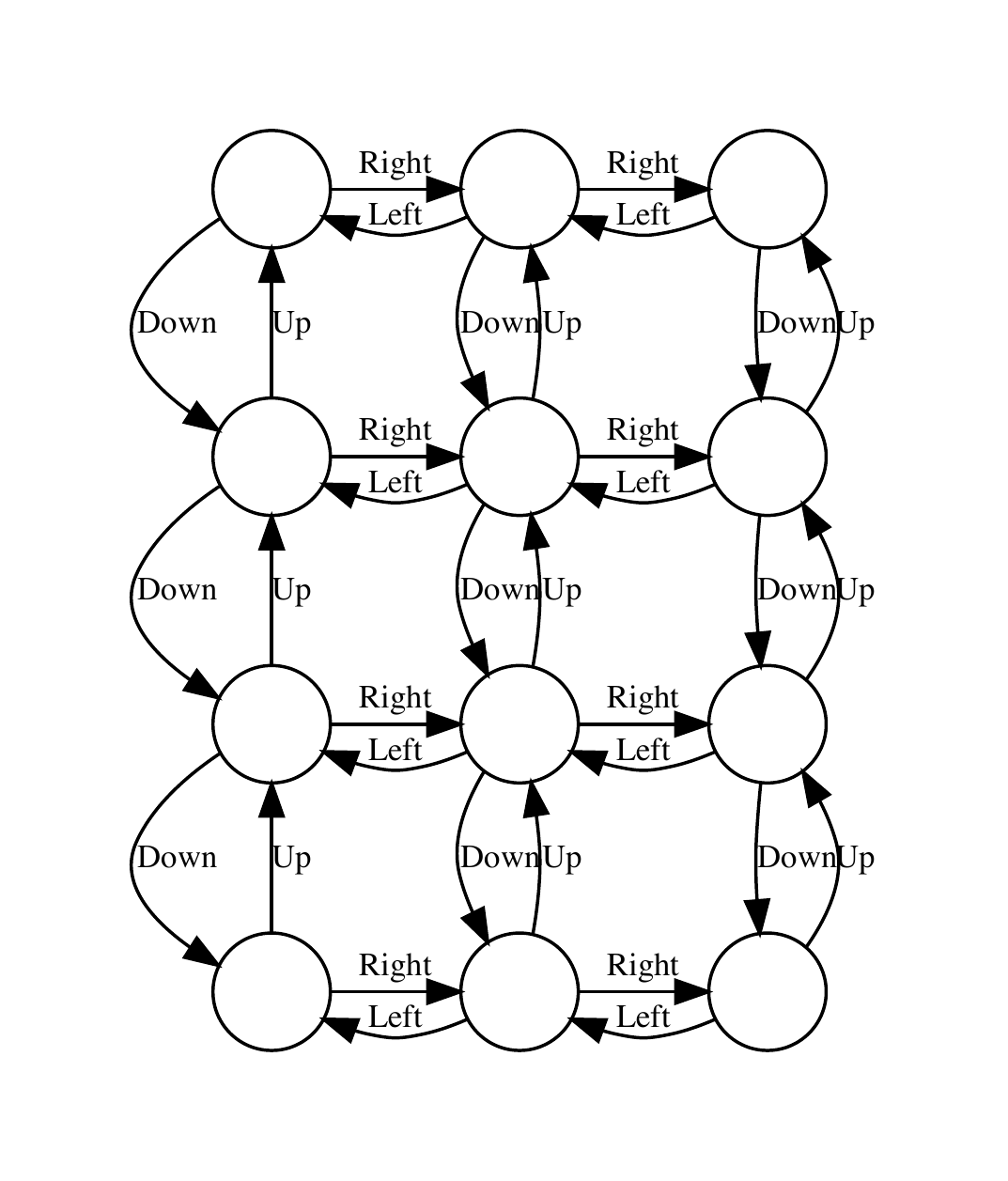}} \\
    {\footnotesize (a) Towers of Hanoi (1 label)} &&
    {\footnotesize (b) Gripper (3 labels)} &&
    {\footnotesize (c) Blocksworld (3 labels)} &&
    {\footnotesize (d) $4{\times}3$ Grid (4 labels)}
  \end{tabular}
  \vskip -.6em
  \caption{Input data for learning the planning representations in the four  domains considered. The proposed
    formulation accepts one or mode labeled directed graphs  encoding the structure of the state space of
    one or more problem instances  as the \emph{sole input}.  It then  produces  symbolic PDDL-like planning
    representations that account for the input graphs in the form of a general first-order domain with
    action schemas and predicate symbols, and instance information describing  objects  and an initial situation.
    Labels are used to distinguish action types.   In each of the four domains, a  single input graph
    corresponding to a single instance  (as shown) sufficed to learn the general first-order domain representations.
    (The graphs can be zoomed in to reveal the labels.)}
  \label{fig:graphs}
\end{figure*}

\subsection{Outputs: First-Order Representations}

Given  labeled graphs $G_1, \ldots, G_m$ in the input,
the learning method produces a corresponding set of planning instances $P_1, \ldots, P_m$
in the output over a common planning domain $D$ (also learned).
A (classical) planning instance is a pair  $P\,{=}\,\tup{D,I}$ where
$D$ is a  \textbf{first-order planning  domain} and $I$ is the \textbf{instance information}.
The  planning domain $D$ contains a set of predicate symbols and a set of  action schemas with preconditions
and effects given by atoms $p(x_1, \ldots, x_k)$ or their negations,  where  $p$ is a domain predicate
and each $x_i$ is a variable representing one of the arguments of the action schema.
The instance information is a  tuple  $I\,{=}\,\tup{O,Init,Goal}$   where $O$ is a (finite) set of object
names $c_i$, and $Init$ and $Goal$ are  sets  of ground atoms $p(c_1, \ldots, c_k)$ or their negations,
where  $p$ is a  predicate symbol in $D$ of arity $k$. This is    the structure of planning problems
expressed in  PDDL \cite{pddl,pddl:book} that corresponds to  STRIPS schemas with negation.
The actual  name of the constants in $O$ is  irrelevant and can be  replaced
by numbers in the interval $[1,N]$ where $N = |O|$ is the number of objects in $O$.
Similarly,  goals are  included in $I$ to keep the notation consistent with
planning practice, but they play no role in the formulation.

\Omit{
Most  current planners take a   planning problem $P$ and for efficiency reasons,
they first  replace   the action schemas   by  their possible instantiations over the constants in $O$.
The  first-order representation of  $D$  is essential  however  for transparency
and for defining  new instances  in terms of new  tuples $I\,{=}\,\tup{O,Init,Goal}$.
}

A problem $P\,{=}\,\tup{D,I}$ defines a \emph{labeled}  graph $G(P)\,{=}\,\tup{V,E,L}$
where the  nodes $n$   in $V$ correspond to the states $s(n)$ over $P$, and there is an
edge  $(n,n')$ in $E$ with label $a$,  $(n,a,n')$, if the state transitions $(s(n),s(n'))$
is enabled by a ground instance of the schema $a$ in $P$.
It is thus assumed that the   ground instances of the same action schema
share the same label, and hence that  edges with different labels in the input graphs
involve  ground instances from different action schemas.

\subsection{Inputs to Outputs: Representation Discovery}

Representation learning in our setting is about finding the (simplest) domain $D$ and instances $P_i$ over $D$
that define graphs $G(P_i)$ that are structurally equivalent (isomorphic) to the input graphs $G_i$.
We formalize the relation between the labeled graphs $G(P_i)$ associated
with the instances $P_i$ and the input labeled graphs $G_i$ as follows:

\begin{definition}
An   instance $P$ \textbf{accounts for} a  labeled graph $G$
if   $(n,a,n')$ is a  labeled edge in  $G(P)$ iff $(h(n),g(a),h(n'))$ is a labeled edge
in $G$, for some function   $g$  between the labels in $G(P)$ and those in $G$,
and a  1-to-1 function $h$   between the nodes in $G(P)$ and those in $G$.
\label{def:account}
\end{definition}

\noindent The representation learning problem is then:

\begin{definition}
The \textbf{representation discovery problem} is finding  a  domain $D$
and instances $P_i\,{=}\,\tup{D,I_i}$ that account for the  input labeled graphs $G_i$, $i=1, \ldots, m$.
\label{def:rdp}
\end{definition}

For solving the problem, we take advantage that  the space of possible domain representations
$D$ is  bounded by the values of a small number of domain  parameters like the number of action schemas,
predicates, and arguments (arities). Likewise, the   number of possible instances $I_i$ is bounded by
the size of the input  graphs.  As a result,  representation discovery becomes a
{combinatorial} problem.  The domain  parameters   define also how complex  a domain representation is,
with simpler representations involving parameters with smaller values.
\emph{Simpler domain representations are preferred}  although we do not introduce or deal with
functions to rank domains. Instead, we simply bound the value of such parameters.


\Omit{
there will not be  just one, but a collection of \textbf{simplest} (Pareto optimal) domains and instances
that follow from  Definition~\ref{def:rdp}. By weighting the values of these parameters, we could
address and solve the problem as a combinatorial optimization problem, although in this work, we will instead
a  SAT formulation that  will look   for solutions  where the max  values of these parameters is bounded.
}

\section{SAT ENCODING}

The problem of computing the instances $P_i\,{=}\,\tup{D,I_i}$ that account for
the observed labeled graphs $G_i$, $i\,{=}\,1, \ldots, n$, is mapped into
the problem of checking the satisfiability of a propositional theory $T_\alpha(G_{1:n})$
where $\alpha$ is a vector of hyperparameters.
The theory $T_\alpha(G_{1:n})$ is  the union of  formulas or layers

\begin{equation}
T_\alpha(G_{1:n}) = T^0_\alpha \, \cup\ \, \bigcup\,\{ T^i_\alpha : i = 1, 2, \ldots, n \}
\end{equation}

\noindent where the  formula   $T^0_\alpha$ is aimed at capturing the domain $D$,
and takes as input the  vector $\alpha$ of hyperparameters only,
while formula  $T^i_\alpha$ is aimed at capturing the instance information $I_i$,
and takes as input the graph $G_i$ as well.  
The \emph{domain layer} $T^0_\alpha$  involves its own variables, while each \emph{instance layer}
$T^i_\alpha$ involves its own variables and those of the domain layer. The encoding of
the domain $D$ and the instances $P_i$ can  be read (decoded) from the   truth assignments
that satisfy the theory  $T_\alpha(G_{1:n})$ over the variables in the corresponding layer.

The \textbf{vector of hyperparameters} $\alpha$  represents the number  of action schemas and the arity of  each  one of them,
the number of predicate symbols  and the arity of each one of them, the number of different atoms in the schemas,
the total  number of  unary and binary static predicates, and the number of objects  in each layer $i$.
We provide a max value  on each of  these parameters, and then consider the theories
$T_\alpha(G_{1:n})$ for each of the   $\alpha$ vectors that comply with  such bounds.
Action schemas $a$, predicate symbols $p$, atom names  $\metaf$, arguments $\metao$, unary and binary predicates $u$ and $b$, and objects $o$,
are all integers that range from $1$ to their corresponding number in $\alpha$. A predicate
$p$ is static if all  $p$-atoms are static, meaning that they do not appear in the effects of any action.
Static atoms are used as preconditions to control the grounding of action schemas, and in the SAT encoding,
they are treated differently than the  other (fluent) atoms.

Next we fully define the theory $T_\alpha(G_{1:n})$: first the domain layer
$T^0_\alpha$ and then each of the instance layers $T^i_\alpha$, $i=1, \ldots, n$. The encoding
is not trivial and it is one of the main contributions of this work, along with the formulation
of the representation learning problem and the results.
For lack of space, we only provide brief explanations for the formulas in the encoding.

\subsection{SAT Encoding: Domain Layer $T^0_\alpha$}

The domain layer $T_\alpha^0$ makes use of the following boolean variables,
some of which can be regarded as  {decision} variables, and the others,
as the  variables whose values  are determined by them. It defines the space
of possible domains $D$ given the value of the hyperparameters $\alpha$
and it does not use the input graphs.

\medskip
\noindent Decision propositions: \\[-1.5em]
\begin{enumerate}[{\small$\bullet$}]
  \item $\PreZero(a,\metaf)/\PreOne(a,\metaf)$: $\metaf$ is negative/positive precondition  of $a$,
  \item $\EffZero(a,\metaf)/\EffOne(a,\metaf)$: $\metaf$ is negative/positive effect of $a$,
  \item $\Label(a,l)$: label of action schema  $a$ is $l$,
  \item $\Arity(p,i)$: arity of predicate symbol $p$, 
  \item $\Atom(\metaf,p)$:  $\metaf$ is   a $p$-atom,
  \item $\Atom(\metaf,i,\metao)$: $i$-th argument of $\metaf$ is (action) argument $\metao$, 
  \item $\Unary(u,a,\metao)$: action $a$ uses static unary predicate $u$ on argument $\metao$,
  \item $\Binary(b,a,\metao,\metao')$: $a$ uses static binary pred.\ $b$ on arguments $\metao$ and $\metao'$.
\end{enumerate}

\noindent Implied propositions: \\[-1.5em]
\begin{enumerate}[{\small$\bullet$}]
  \item $\Using(a,\metaf)$: action $a$ uses atom $\metaf$,
  \item $\Using(\metaf)$: some action $a$ uses atom $\metaf$,
  \item $\Arg(a,\metao)$: action $a$ uses argument $\metao$,
  \item $\Relevant(a,\metao,\metaf,i) \Iff \Using(a,\metaf) \land \Atom(\metaf,i,\metao)$,
  \item $\neg\StaticZero(a,\metaf,p) \Rightarrow \Atom(\metaf,p) \land \EffZero(a,\metaf)$,
  \item $\neg\StaticOne(a,\metaf,p) \Rightarrow \Atom(\metaf,p) \land \EffOne(a,\metaf)$.
\end{enumerate}

\subsubsection{Formulas}
\vskip -1em
\begin{alignat}{1}
  \shortintertext{\Head{Atoms in  preconditions and effects, and unique action labels:}}
  &\Using(a,\metaf){\Iff}\PreZero(a,\metaf){\lor}\PreOne(a,\metaf){\lor}\EffZero(a,\metaf){\lor}\EffOne(a,\metaf) \\
  &\Using(\metaf) \Iff \textstyle\bigvee_a \Using(a,\metaf) \\
  &\neg\PreZero(a,\metaf) \lor \neg\PreOne(a,\metaf) \\
  &\neg\EffZero(a,\metaf) \lor \neg\EffOne(a,\metaf) \\
  &\AMO \{ \Label(a,l) : l \} \\
  \shortintertext{\Head{Effects are non-redundant, and unique  predicate arities:}}
  &\EffZero(a,\metaf) \Implies \neg\PreZero(a,\metaf) \\
  &\EffOne(a,\metaf) \Implies \neg\PreOne(a,\metaf) \\
  &\ExactlyOne \{ \Arity(p,i) : 0 \leq i \leq \MaxAtomArity \} \\
  %
    \shortintertext{\Head{Structure of atoms: predicate symbols and arguments:}}
  &\ExactlyOne \{ \Atom(\metaf,p) : p \} \\
  &\AMO \{ \Atom(\metaf,i,\metao) : \metao \} \\
  &\Atom(\metaf,p) \land \Atom(\metaf,i,\metao) \Implies \textstyle\bigvee_{i \leq j \leq \MaxAtomArity} \Arity(p,j) \\
  &\Atom(\metaf,p) \land \Arity(p,i) \Implies \textstyle\bigwedge_{1\leq j\leq i} \bigvee_{\metao} \Atom(\metaf,j,\metao) \\
  &\Atom(\metaf,p) \land \Arity(p,i) \Implies \textstyle\bigwedge_{i<j\leq\MaxAtomArity} \neg\Atom(\metaf,j,\metao) \\
  %
  \shortintertext{\Head{1-1 map of atom names into possible atom  structures:}
    For $\text{vect}(\metaf)$ denoting  the  boolean  vector with components
    $\Using(\metaf)$, $\{\Atom(\metaf,p)\}_p$, and $\{\Atom(\metaf,i,\metao)\}_{i,\metao}$,
    impose the constraint $\text{vect}(\metaf)\,{<_{lex}}\,\text{vect}(\metaf')$ when $\metaf\,{<}\,\metaf'$
    for uniqueness.}
  %
  &\SLex \{ \text{vect}(\metaf) : \metaf \} \\
  \shortintertext{\Head{Atoms are non-static; static atoms dealt with separately:}}
  &\textstyle\bigvee_{a,\metaf} \bigl[ \neg\StaticZero(a,\metaf,p) \lor \neg\StaticOne(a,\metaf,p) \bigr] \\
  &\neg\StaticZero(a,\metaf,p) \Implies \Atom(\metaf,p) \land \PreOne(a,\metaf) \land \EffZero(a,\metaf) \\
  &\neg\StaticOne(a,\metaf,p) \Implies \Atom(\metaf,p) \land \PreZero(a,\metaf) \land \EffOne(a,\metaf) \\
  %
  \shortintertext{\Head{Atom and action arguments:}}
  &\Using(a,\metaf) \land \Atom(\metaf,i,\metao) \Implies \Arg(a,\metao) \\
  &\Arg(a,\metao) \Implies \textstyle\bigvee_{\metaf,i} \Relevant(a,\metao,\metaf,i) \\
  &\Relevant(a,\metao,\metaf,i) \Iff \Using(a,\metaf) \land \Atom(\metaf,i,\metao) \\
  %
  \shortintertext{\Head{Arities of action schemas and predicate symbols:}}
  &\textstyle\bigwedge_{\metao \geq \ArityOf{action $a$}} \neg\Arg(a,\metao) \\
  &\textstyle\bigwedge_{0\leq\metao<\ArityOf{action $a$}} \Arg(a,\metao) \\
  &\textstyle\bigwedge_{i \neq \ArityOf{atom $p$}} \neg\Arity(p,i) \land \textstyle\bigwedge_{i=\ArityOf{atom $p$}} \Arity(p,i) \\
  \shortintertext{\Head{If static predicate on action argument, argument must exist:}}
  &\Unary(u,a,\metao) \Implies \Arg(a,\metao) \\
  &\Binary(b,a,\metao,\metao') \Implies \Arg(a,\metao) \land \Arg(a,\metao')
\end{alignat}

\subsection{SAT Encoding: Instance Layer $T^i_\alpha$}


The layers $T^i_\alpha$ of the propositional theory $T_\alpha(G_{1:n})$
make use of the input graphs $G_i$ in the form of a set of
states (nodes) $s$ and transitions (edges) $t$.
The source and destination states of a transition
$t$ are denoted $t.src$ and $t.dest$, and the label
as  $t.label$. The layers $T^i_\alpha$ introduce symbols
for ground atoms $k$, objects $o$, and tuples of  objects  $\bar o$
whose size matches the arity of the context where they are
used (action and predicate arguments). The number of ground
atoms $k$ is determined by the number of objects,
predicate symbols,  and arguments, as established
by the hyperparameters in $\alpha$. The index $i$ that refers
to the $i$-th input graph $G_i$ is omitted for readability.

\medskip
\noindent Decision propositions:
\begin{enumerate}[{\small$\bullet$}]
  \item $\Map(t,a)$: transition $t$ is mapped to action schema $a$,
  \item $\Mapf(t,k,\metaf)$: ground atom $k$  is mapped to  atom $\metaf$ in transition $t$,
  \item $\phi(k,s)$: value of (boolean) ground atom $k$ at state $s$,
  \item $\Ground(k,p)$: ground atom $k$ refers to predicate symbol  $p$,
  \item $\Ground(k,i,o)$: $i$-th argument of ground atom  $k$ is object $o$ [$i>0$],
  \item $r(u,o)$: true if $u(o)$ holds for  static unary predicate $u$,
  \item $s(b,o,o')$: true if $b(o,o')$ holds for static binary predicate $b$,
  \item $\Gtuple(a,\bar{o})$: true if $a(\bar{o})$ is a ground instance of $a$.
\end{enumerate}

\noindent Implied propositions:
\begin{enumerate}[{\small$\bullet$}]
  \item $\Free(k,t,a)$: ground atom $k$ is unaffected in trans.\ $t$ mapped to $a$,
  \item $\G(k,s,s') \Iff \phi(k,s) \oplus \phi(k,s')$ ($\oplus$ is XOR),
  \item $U(u,a,\metao,o) \Iff \Unary(u,a,\metao) \land \neg r(u,o)$,
  \item $B(b,a,\metao,\metao',o,o') \Iff \Binary(b,a,\metao,\metao'){\land}\neg s(b,o,o')$,
  \item $\Mapt(t,\metao,o)$: argument $\metao$ is mapped to object $o$ in transition $t$,
  \item $W(t,k,i,\metao) \Implies \bigl[ \Ground(k,i,o) \Leftrightarrow \Mapt(t,\metao,o) \bigr]$,
  \item $G(t,a,\bar{o})$: transition $t$ is (ground) instance of $a(\bar{o})$,
  \item $\Appl(a,\bar{o},s)$: ground instance $a(\bar{o})$ is applicable in state $s$,
  \item $\ViolatedZero(a,\bar{o},s,k)$:   $k$ is neg.\ precond.\ $p(\bar{o})$  of $a$ that is false in $s$,
  \item $\ViolatedOne(a,\bar{o},s,k)$:   $k$ is pos.\  precond.\ $p(\bar{o})$  of $a$ that is false in $s$,
  \item $\PreZeroEq(a,\bar{o},k,\metaf) \Rightarrow \PreZero(a,\metaf) \land \Eq(\bar{o},\metaf,k)$,
  \item $\PreOneEq(a,\bar{o},k,\metaf) \Rightarrow \PreOne(a,\metaf) \land \Eq(\bar{o},\metaf,k)$,
  \item $\Eq(\bar{o},\metaf,k)$: ground atom $k$ instantiates  atom  $\metaf$ with tuple $\bar{o}$.
\end{enumerate}

\subsubsection{Formulas}
\vskip -1em
\begin{alignat}{1}
  \shortintertext{\Head{Binding transitions with action schemas,  and ground atoms with atom  schemas:}}
  &\ExactlyOne \{ \Map(t,a) : a \} \\
  %
  &\AMO \{ \Mapf(t,k,\metaf) : \metaf \} \\
  &\AMO \{ \Mapf(t,k,\metaf) : k \} \\
  \shortintertext{\Head{Consistency between mappings, labeling, and usage:}}
  &\Map(t,a) \Implies \Label(a,t.label) \\
  &\Map(t,a) \land \Mapf(t,k,\metaf) \Implies \Using(a,\metaf) \\
  &\Map(t,a) \land \Using(a,\metaf) \Implies \textstyle\bigvee_k \Mapf(t,k,\metaf) \\
  %
  \shortintertext{\Head{Ground atom unaffected if:}}
  &\Map(t,a) \land \bigl[ \textstyle\bigwedge_{\metaf} \neg\Mapf(t,k,\metaf) \bigr]  \Implies \Free(k,t,a) \\
  &\notag\Map(t,a) \land \Mapf(t,k,\metaf) \\
  &\qquad\Implies \bigl[ \neg\EffZero(a,\metaf) \land \neg\EffOne(a,\metaf) \Iff \Free(k,t,a) \bigr] \\
  \shortintertext{\Head{Transitions and inertia:}}
  &\Map(t,a) \land \Mapf(t,k,\metaf) \land \PreZero(a,\metaf) \Implies \neg\phi(k,t.src) \\
  &\Map(t,a) \land \Mapf(t,k,\metaf) \land \PreOne(a,\metaf) \Implies \phi(k,t.src) \\
  &\Map(t,a) \land \Mapf(t,k,\metaf) \land \EffZero(a,\metaf) \Implies \neg\phi(k,t.dst) \\
  &\Map(t,a) \land \Mapf(t,k,\metaf) \land \EffOne(a,\metaf) \Implies \phi(k,t.dst) \\
  &\Map(t,a){\Implies}\bigl[ \Free(k,t,a){\Iff} [ \phi(k,t.src){\Iff}\phi(k,t.dst) ] \bigr] \\
  %
  \intertext{\Head{States must differ in value of some ground atom:}}
  &\G(k,s,s') \Iff \phi(k,s) \oplus \phi(k,s') \\
  &\textstyle\bigwedge_{s<s'} \textstyle\bigvee_{k} \G(k,s,s') \\ 
  \shortintertext{\Head{Predicate symbol and arguments of  ground atoms:}}
  &\ExactlyOne \{ \Ground(k,p) : p \} \\
  &\AMO \{ \Ground(k,i,o) : o \} \\
  &\Ground(k,p) \land \Ground(k,i,o) \Implies \textstyle\bigvee_{i \leq j \leq \MaxAtomArity} \Arity(p,j) \\
  &\Ground(k,p) \land \Arity(p,i) \Implies \textstyle\bigwedge_{1 \leq j \leq i} \textstyle\bigvee_{o} \Ground(k,j,o) \\ 
  &\Ground(k,p) \land \Arity(p,i) \Implies \textstyle\bigwedge_{i < j} \neg\Ground(k,j,o) \\ 
  %
  \shortintertext{\Head{1-1 map of ground atoms names to predicates and arguments:}
    For $\text{vect}(k)$ denoting boolean  vector with components $\{\Ground(k,p)\}_p$ and $\{\Ground(k,i,o)\}_{i,o}$,
    impose constraint $\text{vect}(k)\,{<_{lex}}\,\text{vect}(k')$ for $k\,{<}\,k'$.}
%
%
  &\SLex \{ \text{vect}(k) : k \} \\
  \shortintertext{\Head{Ground atoms  and schema atoms in sync:}}
  &\Mapf(t,k,\metaf) \Implies \bigl[ \Atom(\metaf,p) \Iff \Ground(k,p) \bigr] \\
  &\Mapf(t,k,\metaf) \land \Atom(\metaf,i,\metao) \Implies \textstyle\bigvee_{o} \Ground(k,i,o) \\
  &\Mapf(t,k,\metaf) \land \Ground(k,i,o) \Implies \textstyle\bigvee_{\metao} \Atom(\metaf,i,\metao) \\
  \shortintertext{\Head{Excluded bindings  of static predicates:}}
  &U(u,a,\metao,o) \Iff \Unary(u,a,\metao) \land \neg r(u,o)  \\
  &B(b,a,\metao,\metao',o,o') \Iff \Binary(b,a,\metao,\metao'){\land} \neg s(b,o,o') \\
  \shortintertext{\Head{Bindings associated with transitions (part 1):}}
  &\AMO \{ \Mapt(t,\metao,o) : o \} \\
  &\Map(t,a) \land \Arg(a,\metao) \Implies \textstyle\bigvee_{o} \Mapt(t,\metao,o) \\
  &\Map(t,a) \land \Mapt(t,\metao,o) \Implies \Arg(a,\metao) \\
  \shortintertext{\Head{Bindings associated with transitions (part 2):}}
  &\Mapf(t,k,\metaf) \land \Atom(\metaf,i,\metao) \Implies W(t,k,i,\metao) \\
  &W(t,k,i,\metao) \Implies \bigl[ \Ground(k,i,o) \Iff \Mapt(t,\metao,o) \bigr] \\
  %
  %
  \shortintertext{\Head{Explanation of non-existing ground actions $\Gtuple(a,\bar{o})$:}}
  &\notag\neg\Gtuple(a,\bar{o}) \Implies \textstyle\bigvee_{o_i>0} \neg\Arg(a,\metao_i)\,\lor\, \textstyle\bigvee_{u,i} U(u,a,\metao_i,o_i)\ \lor \\
  \label{eq:explanation-non-existent-ground-actions}
  &      \qquad\qquad\qquad\ \ \ \,\textstyle\bigvee_{b,i<j} B(b,a,\metao_i,\metao_j,o_i,o_j) \\
  \intertext{\Head{Explanation of existing ground actions:}}
  &\Map(t,a) \land \Mapt(t,\metao,o) \land \Unary(u,a,\metao) \Implies r(u,o) \\
  &\Map(t,a) \land \Mapt(t,\metao,o) \land \Mapt(t,\metao',o') \land \Binary(b,a,\metao,\metao') \\
  &\qquad\qquad\quad\ \, \Implies s(b,o,o') \\
  %
  %
  \intertext{\Head{Ground actions must be used in some transition:}}
  \label{eq:ground-actions-must-be-used}
  &\Gtuple(a,\bar{o}) \Implies \textstyle\bigvee_{ t} G(t,a,\bar{o}) \\
  &G(t,a,\bar{o}) \Implies \Gtuple(a,\bar{o}) \\
  &\notag G(t,a,\bar{o}) \Implies \Map(t,a)\ \land\ \textstyle\bigwedge_{o_i>0} \Mapt(t,\metao_i,o_i)\ \land \\
  &\qquad\qquad\quad\ \, \textstyle\bigwedge_{o_i=0} \bigl[ \Arg(a,\metao_i) \Implies \Mapt(t,\metao_i,o_i) \bigr] \\
  &\AMO \{ G(t,a,\bar{o}) : t.src = s \} \\ 
  &\ExactlyOne \{ G(t,a,\bar{o}) : a, \bar{o} \} \\
  \intertext{\Head{Applicable actions must be applied:}}
  &G(t,a,\bar{o}) \Implies \Appl(a,\bar{o},t.src) \\
  &\Appl(a,\bar{o},s) \Implies \textstyle\bigvee_{t.src = s} G(t,a,\bar{o}) \\
  &\notag\neg\Appl(a,\bar{o},s) \Implies \neg\Gtuple(a,\bar{o})\ \lor \\
  &\qquad\qquad\qquad\quad\ \textstyle\bigvee_k \bigl[ \ViolatedZero(a,\bar{o},s,k) \lor \ViolatedOne(a,\bar{o},s,k) \bigr] \\
  &\ViolatedZero(a,\bar{o},s,k) \Implies \phi(k,s) \land \textstyle\bigvee_{\metaf} \PreZeroEq(a,\bar{o},k,\metaf) \\
  &\ViolatedOne(a,\bar{o},s,k) \Implies \neg\phi(k,s) \land \textstyle\bigvee_{\metaf} \PreOneEq(a,\bar{o},k,\metaf) \\
  &\PreZeroEq(a,\bar{o},k,\metaf) \Implies \PreZero(a,\metaf) \land \Eq(\bar{o},\metaf,k) \\
  &\PreOneEq(a,\bar{o},k,\metaf) \Implies \PreOne(a,\metaf) \land \Eq(\bar{o},\metaf,k) \\
  &\Eq(\bar{o},\metaf,k) \Implies \bigl[ \Atom(\metaf,p) \Iff \Ground(k,p) \bigr] \\
  &\Eq(\bar{o},\metaf,k) \land \Atom(\metaf,i,\metao_j) \Implies \Ground(k,i,o_j) 
\end{alignat}

The encoding also contains formulas that reduce the  number of redundant, symmetric valuations, which  are omitted here for clarity.
Such formulas only affect the performance of SAT solvers and do not affect the satisfiability of the theory $T_\alpha(G_{1:n})$.


\Omit{
\subsection{Symmetries}

Formulas that reduce the number of redundant, symmetric, valuations of the  theory.
\begin{alignat}{1}
  \intertext{\Head{(Symmetry) Ordered arities and arguments:}}
  &\Arity(1+p,i) \Implies \textstyle\bigvee_{0 \leq j \leq i} \Arity(p,j) \\
  &\Arg(a,1+\metao) \Implies \Arg(a,\metao) \quad \\
  \intertext{\Head{(Symmetry) Definition of $\Ord(o,k,i,s)$:}}
  &\Ord(o,k,i,s) \Iff \Ground(k,i,o) \land \phi(k,s) \\
  \intertext{\Head{(Symmetry) Ordered objects at each layer:
  vector for object $o$ is: $\tup{\Ord(o,k,i,s) : k, i, s}$ in order $s$, $k$, $i$}}
  &\tup{\text{non-strict-lexicographic-ordering-objects-in-layers}}
  \intertext{\text{\textcolor{red}{NOTE:} It is still unclear whether this formula preserves satisfiability}}
\end{alignat}
}

\section{PROPERTIES}

The correctness and completeness of the encoding is expressed as:  

\begin{thm}
  The  instances $P_i\,{=}\,\tup{D,I_i}$ with parametrization $\alpha$  account for the input labeled graphs 
  $G_1, \ldots, G_n$  applying every  ground action  at least once iff there is a satisfying assignment
  of the theory $T_{\alpha}(G_{1:n})$ that encodes these instances up to renaming.
\end{thm}

This means basically that if the graphs can be generated by some instances, such instances are encoded in one of the
models of the SAT encoding. On the other hand, any satisfying assignment of the theory encodes a first-order
domain $D$ and instances $P_i$ over $D$ that solve the representation discovery problem for the input graphs.

The parametrization $\alpha$ associated with a set of instances with a shared domain
is simply the value of the hyperparameters determined by the instances.
The condition that ground actions must be applied at least once follows from \eqref{eq:ground-actions-must-be-used}
and could be relaxed.  In the encoding, indeed, if a  ground action $a(\bar o)$
is never applied (i.e., $\Gtuple(a,\bar{o})$ is false), it must be because the static predicates filter it out
(cf.\ second and third disjunctions in \eqref{eq:explanation-non-existent-ground-actions}).
On the other hand, the first disjunct in \eqref{eq:explanation-non-existent-ground-actions} explains inexistent ground
actions due to ``wrong groundings'';
namely, groundings of variables that are not arguments of the action schema. 

The extraction of instances  $P_i\,{=}\,\tup{D,I_i}$, $I_i\,{=}\,\tup{O_i,Init_i,Goal_i}$,
from a satisfying assignment is direct for the domain $D$ and the objects $O_i$ in each instance $I_i$. 
The assignment embeds each node $n$ of the input graph $G_i$ into a first-order state $s(n)$ over the problem $P_i$.
The initial state $Init_i$ can be set to any state $s(n)$ for a node $n$ in $G_i$ that is connected to all other nodes in $G_i$, 
while $Goal_i$ plays no role in the structure of the state space and it is left
unconstrained. 

Finally, observe that the size of the theory is exponential only in the hyperparameter that specifies
the max arity of action schemas since the tuples $\bar o$ of objects that define grounded actions $a(\bar o)$
appear explicit in the formulas. However, the arities of action schemas are bounded  and  small; we use
a bound of 3 in all the experiments.

\section{VERIFICATION}

It is possible to verify the representations learned by leaving apart some input graphs $G_k$, $k > n$, for
\textbf{testing only} as it is standard in supervised  learning. For this, the learned domain $D$
is verified with respect to each testing graph $G_k$ individually, by checking whether there is an
instance $P_k\,{=}\,\tup{D,I_k}$ of the learned  domain $D$ that accounts for the graph $G_k$, following Def.~\ref{def:account}.
This test may   be also performed with a SAT solver over a propositional theory  $T'(G_k)$  that is a simplified version
of the theory $T_\alpha(G_{1:n})$.  Indeed, if the domain $D$ was obtained from
a satisfying truth assignment $\sigma$ for the theory $T_\alpha(G_{1:n}) = T_\alpha^0 \cup \cup_{i=1,n} T_\alpha^i$,
then $T'(G_k) = T_{\alpha}^0 \cup T_\alpha^k \cup \sigma^0$ where $\sigma^0$ is the set of literals
that captures the valuation $\sigma$ over the symbols in $T_\alpha^0$. In words, $T'(G_k)$ treats $G_k$
as an input graph but with the values of the  domain  literals in layer $T_\alpha^0$  set to the values
in $\sigma^0$.

\begin{table*}
  \centering
  \caption{Instance, \# of labels,  nodes and edges in graph,
    \# of parametrizations $\alpha$ and  theories $T_\alpha(G)$, fraction evaluated, and \#  found to be indeterminate (SAT solver
    still running after 1h cutoff), UNSAT, or SAT, with $x+y+z$ meaning  that $x$ did not complete verification  in time/memory  bound,
    $y$ failed it, and  $z$ passed it (solutions). Last columns show avg.\ sizes and times of theories that produced these solutions.
  }
  \resizebox{\textwidth}{!}{
    \begin{tabular}{l rrr c rrrrr c rrrr}
                          & \multicolumn{3}{c}{Input TS} && \multicolumn{5}{c}{Statistics SAT Calls} && \multicolumn{4}{c}{Theory for SAT Tasks (avg.)} \\
      \cmidrule{2-4}\cmidrule{6-10}\cmidrule{12-15}
      Instance                   & \#labels & \#states & \#trans. && \#tasks & sample & INDET & UNSAT &        SAT &&       \#vars &    \#clauses &    time & mem.\ (Mb) \\
      \midrule
      Blocksworld (4blocks)              & 3 &       73 &      240 &&  19,050 &  1,905 &   246 & 1,642 &    10+0+7 &&  1,666,705.5 &  6,033,529.0 & 1,441.1 &      860.7 \\ 
      Towers of Hanoi (3disks + 3pegs)   & 1 &       27 &       78 &&   6,390 &    639 &    24 &   614 &     0+0+1 &&    860,704.0 &  3,328,492.0 & 1,691.7 &      454.5 \\ 
      Gripper (2rooms + 3balls)          & 3 &       88 &      280 &&  19,050 &  1,905 &   333 & 1,564 &     0+2+6 &&  1,592,358.5 &  6,176,073.3 & 1,840.3 &      873.4 \\ 
      Rectangular grid $4{\times}3$      & 4 &       12 &       34 &&  37,800 &  3,780 &    55 & 3,496 & 10+141+78 &&    321,904.0 &  1,165,860.6 &   156.7 &      164.5 \\ 
      Rectangular grid $4{\times}3$      & 2 &       12 &       34 &&  15,120 &  1,512 &    36 & 1,408 &    2+4+62 &&    343,472.7 &  1,299,706.1 &    46.3 &      175.4 \\ 
      Rectangular grid $4{\times}3$      & 1 &       12 &       34 &&   7,560 &    756 &    11 &   715 &    2+0+28 &&    363,418.2 &  1,683,392.7 &    53.4 &      211.0 \\ 
      \bottomrule
    \end{tabular}
  }
  \label{table:exp}
 \end{table*}

\section{EXPERIMENTS AND RESULTS}

We performed experiments to test the computational feasibility of the approach and the type of first-order representations that are  obtained.
We considered four domains, Blocksworld, Towers of Hanoi, Grid, and Gripper.  For  each domain, we selected a single input graph $G=G_1$
of  a small instance to build the theory $T_\alpha(G_{1:n})$ with $n=1$, abbreviated  $T_\alpha(G)$,
converted it to  CNF, and fed it to the  SAT solver \texttt{glucose-4.1} \cite{glucose}.
The input graphs used in the experiments are shown in Fig.\,\ref{fig:graphs}.
The experiments were performed on Amazon EC2's  \texttt{c5n.18xlarge} with a limit of 1 hour and 16Gb of memory.
If $T_\alpha(G)$, for parameters $\alpha$  was found to be  satisfiable, we obtained  an  instance $P\,{=}\,\tup{D,I}$.
The size of these graphs
in terms of the number of nodes and edges appear in Table~\ref{table:exp} as 
\#states and \#trans, while
\#tasks  is the number of possible  parametrizations $\alpha$ that results from the following bounds:
\begin{enumerate}[--]
  \item max number of action schemas set to number of labels,
  \item max number of predicate symbols set to 5,
  \item max arity of  action schemas and predicates set to 3 and 2 resp, 
  \item max number of atoms schemas set to 6,\footnote{The atom schemas are of the form $p(t)$ where $p$ is a predicate symbol
    and $t$ is a tuple of numbers of the arity of $p$, with the numbers representing
    action schema arguments.}
  \item max number of static predicates set to 5,
  \item max number of objects in an instance set to 7.
\end{enumerate}
The choice  for these bounds  is   arbitrary, yet for most benchmarks the first five domain parameters
do not go much higher, and the  last one is  compatible with the idea of learning from  small examples.

\Omit{
\begin{figure*}[t]
  \centering
  \begin{tabular}{ccccccc}
    \resizebox{!}{.45\columnwidth}{\includegraphics{images/blocks2_4}} &&
    \resizebox{.42\columnwidth}{!}{\includegraphics{images/hanoi_3x3}} &&
    \resizebox{.42\columnwidth}{!}{\includegraphics{images/gballs_3x0}} &&
    \resizebox{!}{.45\columnwidth}{\includegraphics{images/grid_3x4_v2}} \\
    {\footnotesize (a) Blocksworld (3 labels)} &&
    {\footnotesize (b) Towers of Hanoi (1 label)} &&
    {\footnotesize (c) Gripper (3 labels)} &&
    {\footnotesize (d) $4{\times}3$ Grid (4 labels)}
  \end{tabular}
  \caption{The input (graph) used for learning each one of the domains.}
  \label{fig:graphs}
\end{figure*}
}

The hyperparameter vector $\alpha$   specifies  the  \emph{exact} values of the parameters,
compatible with the bounds, and the \emph{exact} arities of each action schema and predicate.
This is why there are so many  parametrizations   $\alpha$ and theories $T_\alpha(G)$
to consider (column \#tasks). Given our computational resources, for each input, we run the
SAT solver on 10\% of them randomly chosen. The number of  theories that  are SAT, UNSAT, or
INDET (SAT solver  still running  after   time/memory limit)  are shown in the table
that  also displays  the number of solutions verified on the test  instances.
The last columns show the average sizes of the SAT  theories $T_{\alpha}(G)$ that were solved and verified.
For each domain, we chose one solution at random and display it, with the names of predicates
and action schemas changed to reflect their meanings (i.e., our interpretation).
These solutions are compatible with the hyperparameters but are not necessarily ``simplest'',
as we have not attempted to rank the solutions found.


\subsection{Towers of Hanoi}

The input graph $G$  is the  transition system for Hanoi  with 3 disks, 3 pegs,
and one action  label shown in Fig.\,\ref{fig:graphs}(a).
Only one sampled parametrization $\alpha$ yields
a satisfiable theory $T_\alpha(G)$, and the resulting domain   passes validation
on two test instances, one with 4 disks  and 3 pegs;  the other  with 3 disks and 4 pegs.
%
%
This  solution was  found in 1,692 seconds and uses two
predicates, \atom|clear(d)| and \atom|Non(x,y)|, to indicate that disk $d$ is clear
and that disk $x$ is \emph{not} on disk $y$ respectively.  Two binary static predicates are learned as well,
\atom|BIGGER| and \atom|NEQ|.  The  encoding is correct and intuitive
although it features  negated predicates like \atom|Non| and redundant preconditions
like \atom|Non(fr,d)| and \atom|Non(d,fr)|.
Still, it is remarkable that this subtle first-order  encoding  is obtained from the graph of
one instance, and   that it works  for any instance involving any number of pegs and disks.

\begin{Verbatim}[label=Hanoi (ref. 530)]
!experiments/exp10p-exact-v2/hanoi2/exp10p_enc4t2_hanoi2_n530/theory.cnf 860704 3328492 7.271393 1691.700452 454584 True experiments/exp10p-exact-v2/hanoi2/exp10p_enc4t2_hanoi2_n530/vtheory_ON_7_hanoi_3x4.cnf 4479042 14541438 22.555918 1833.369343 2359296 True experiments/exp10p-exact-v2/hanoi2/exp10p_enc4t2_hanoi2_n530/vtheory_ON_7_hanoi_4x3.cnf 3914115 15274477 22.310272 33.2147 2359296 True
!hanoi2/exp10p_enc4t2_hanoi2_n530
!disks={o1,o3,o4}, pegs={o0,o2,o5}
!p0(?d)=clear(?d), p1(?d0,?d1)=NOT-on(?d0,?d1)
!b0(?d0,?d1)=BIGGER(?d0,?d1), b1(?d0,?d1)=NEQ*(?d0,?d1) {includes (o1,o1)}
!d={o1,o3,o4}, from={o0,o2,o3,o4,o5}, to={o0,o2,o3,o4,o5}
!ordered: o1, o3, o4
\ActionSchema{Move(fr,to,d):}
 Static: BIGGER(fr,d),BIGGER(to,d) NEQ(fr,to)
 Pre: -clear(fr),clear(to),clear(d),Non(fr,d),-Non(d,fr),Non(d,to)
 Eff: clear(fr),-clear(to),Non(d,fr),-Non(d,to)
\end{Verbatim}

\subsection{Gripper}

The instance used to generate the graph $G$ in Fig.\,\ref{fig:graphs}(b) involves  2 rooms, 3 named balls, 2 grippers,
and 3 action labels  for  moves, picks, and drops. In this case, 8 encodings are found,
6 of which pass verification  over instances  with 2 and 4 balls.
One of these encodings, randomly chosen from these 6 is shown below.
It was found in 863 seconds, and uses the atoms  \atom|at(room)|, \atom|hold(gripper,ball)|,
\atom|Nfree(gripper)|, and \atom|Nat(room,ball|) to  denote the  robot position,
that \atom|gripper| holds \atom|ball|, that  \atom|gripper| holds some ball, and
that \atom|ball| is not in \atom|room| respectively.
The learned static predicates are both binary, \atom|CONN| and \atom|PAIR|:
the first for different rooms, and the second, for a pair formed by a  room and  a gripper.
There also redundant preconditions, but the encoding is correct for any number
of rooms, grippers, and balls.

\begin{Verbatim}[label=Gripper (ref. 13918)]
!experiments/exp10p-exact-v2/gballs/exp10p_enc4t2_gballs_n13918/theory.cnf 911922 4121488 9.033631 862.437387 525148 True experiments/exp10p-exact-v2/gballs/exp10p_enc4t2_gballs_n13918/vtheory_ON_2_gripper_balls_2x0.cnf 67579 423975 0.78202 1.216145 525148 True experiments/exp10p-exact-v2/gballs/exp10p_enc4t2_gballs_n13918/vtheory_ON_4_gripper_balls_4x0.cnf 8871627 30292847 69.918433 42.406572 4479664 True
!gballs/exp10p_enc4t2_gballs_n13918
!p0(?loc)=at(?loc), p1(?g)=NOT-free(?g), p2(?g,?ball)=holding(?g,?ball), p1(?loc,?ball)=NOT-at(?loc,?ball)
!b0(?x,y)=pair(?loc,?g), b1(?x,?y)=CONN(?x,?y)
\ActionSchema{Move(from,to):}
 Static: CONN(from,to)
 Pre: at(from),-at(to)
 Eff: -at(from),at(to)

\ActionSchema{Drop(ball,room,gripper):}
 Static: PAIR(room,gripper)
 Pre: at(room),Nfree(gripper),hold(gripper,ball),Nat(room,ball)
 Eff: -Nfree(gripper),-hold(gripper,ball),-Nat(room,ball)

\ActionSchema{Pick(ball,room,gripper):}
 Static: PAIR(room,gripper)
 Pre: at(room),-Nfree(gripper),-hold(gripper,ball),-Nat(room,ball)
 Eff: Nfree(gripper),hold(gripper,ball),Nat(room,ball)
\end{Verbatim}

\subsection{Blocksworld}

The instance used to generate the graph in Fig.\,\ref{fig:graphs}(c) has 4 blocks and 3 action labels
to indicate moves to and from the table, and moves among blocks.
17 of the 10\% of sampled tasks were SAT, and 7 of them complied with test instances
with 2, 3 and 5 blocks. One of these encodings, selected randomly and found in 110 seconds,
is shown below. It has the predicates \atom|Nclear(x)| that holds when (block) $x$ is not clear, and
\atom|Ntable-OR-Non(x,y)|  that holds when $x$ is not on the table for $x\,{=}\,y$, and when block $x$
is on block $y$ for $x\,{\neq}\,y$.  The standard human-written encoding  for Blocksworld features three
predicates instead (\atom|clear|, \atom|ontable|, and \atom|on|). This encoding uses one less predicate but it is more
complex due to the disjunction in \atom|Ntable-OR-Non(x,y)|. As before, some of the preconditions in the schemas are redundant,
and for the action schema \atom|MoveFromTable| the argument $d$ is redundant.


\begin{Verbatim}[label=Blocksworld (ref. 1688)]
!experiments/exp10p-exact-v2/blocks2/exp10p_enc4t2_blocks2_n1688/theory.cnf 1067245 4065351 6.183907 109.134002 559872 True experiments/exp10p-exact-v2/blocks2/exp10p_enc4t2_blocks2_n1688/vtheory_ON_2_blocks2_2.cnf 5013 20974 0.057539 0.045897 559872 True experiments/exp10p-exact-v2/blocks2/exp10p_enc4t2_blocks2_n1688/vtheory_ON_3_blocks2_3.cnf 60858 269419 0.590397 0.607369 559872 True experiments/exp10p-exact-v2/blocks2/exp10p_enc4t2_blocks2_n1688/vtheory_ON_5_blocks2_5.cnf 26282250 74868019 195.380621 116.898943 11934204 True
!blocks2/exp10p_enc4t2_blocks2_n1688
!p0(?x)=NOT-clear(?x), p1(?x,?y)=NOT-ontable-OR-NOT-on(?x,?y)
!b0(?x,?y)=NEQ(?x,?y), b1(?x,?y)=EQ(?x,?y), b2(?x,?y)=NEQ(?x,?y)
\ActionSchema{MoveToTable(x,y):}
 Static: NEQ(x,y)
 Pre: -Nclear(x),Nclear(y),-Ntable-OR-Non(x,y),Ntable-OR-Non(x,x)
 Eff: -Nclear(y),-Ntable-OR-Non(x,x),Ntable-OR-Non(x,y)

\ActionSchema{MoveFromTable(x,y,d):}
 Static: NEQ(x,y),EQ(y,d)
 Pre: -Nclear(x),-Nclear(d),-Ntable-OR-Non(x,x),Ntable-OR-Non(x,y)
 Eff: Nclear(d),Ntable-OR-Non(x,x),-Ntable-OR-Non(x,y)

\ActionSchema{Move(x,z,y):}
 Static: NEQ(x,z),NEQ(z,y),NEQ(x,y)
 Pre: -Nclear(x),Nclear(y),-Nclear(z),Ntable-OR-Non(x,x),
      Ntable-OR-Non(x,z),-Ntable-OR-Non(x,y)
 Eff: Nclear(z),-Nclear(y),Ntable-OR-Non(x,y),-Ntable-OR-Non(x,z)
\end{Verbatim}

\subsection{Grid}

The graph $G$ in Fig.\,\ref{fig:graphs}(d) is for an agent that moves in a  $4{\times}3$ rectangular grid
using three classes of labels:  (the default shown) 4 labels \atom|Up|, \atom|Right|, \atom|Down|, and \atom|Left|,
2 labels \atom|Horiz| and \atom|Vert|, and a unique label \atom|Move|.
Many solutions exist in this problem because the domain is  very simple, even though the space of hyperparameters is   the same.
The randomly chosen  solution  for the  input  with four labels is
complex and it  is not shown. Instead, a simpler and more
intuitive hand-picked solution (found in 3 seconds) is displayed, where
the $x$ position is encoded as usual (one object per position),
but  the $y$ position is encoded with a unary counter (count is number of bits that are  on).

\begin{Verbatim}[label=Grid with 4 labels (ref. 4853)]
!experiments/exp10p-exact-v2/grid/exp10p_enc4t2_grid_n4853/theory.cnf 36074 166748 0.323549 2.300982 25164 True experiments/exp10p-exact-v2/grid/exp10p_enc4t2_grid_n4853/vtheory_ON_5_grid_4x4.cnf 76693 328880 0.739918 0.631118 48524 True experiments/exp10p-exact-v2/grid/exp10p_enc4t2_grid_n4853/vtheory_ON_6_grid_5x6.cnf 217334 874667 1.801575 2.120113 125428 True
!grid/exp10p_enc4t2_grid_n4853
!p0=NatX, p1=NatY
!---------
! Up(o3,o1): -p1(3) => p1(3), -p1(1)
! Up(o1,o0): -p1(1) => p1(1), -p1(0)
! Up(o0,o2): -p1(0) => p1(0), -p1(o2)
!---------
! Down(o0,o2): p1(0), -p1(2) => -p1(0), p1(2)
! Down(o1,o0): p1(1), -p1(0) => -p1(1). p1(0)
! Down(o3,o1): p1(3), -p1(1) => -p1(3), p1(1)
!---------
! Right(o0,o2): p0(0), -p0(2) => p0(2)
! Right(o2,o3): p0(2), -p0(3) => p0(3)
!---------
! Left(o2,o3): p0(2), -p0(3) => -p0(2)
! Left(o3,o1): p0(3), -p0(1) => -p0(3)
!---------
! u0: 0 1 3
! u1: 0 2 3
! b0: (1,0) (0,2) (2,3) (3,1)
!---------
! s0{p0(o0),p1(o0),p1(o1),p1(o2)}: appl={Up(o3,o1)/t0,Right(o0,o2)/t1}
! s1{p0(o0),p1(o0),p1(o2),p1(o3)}: appl={Up(o1,o0)/t2,Right(o0,o2)/t3,Down(o3,o1)/t4}
! s2{p0(o0),p1(o1),p1(o2),p1(o3)}: appl={Up(o0,o2)/t5,Right(o0,o2)/t6,Down(o1,o0)/t7}
! s3{p0(o0),p1(o0),p1(o1),p1(o3)}: appl={Right(o0,o2)/t8,Down(o0,o2)/t9}
! s4{p0(o0),p0(o2),p1(o0),p1(o1),p1(o2)}: appl={Up(o3,o1)/t10,Right(o2,o3)/t11,Left(o2,o3)/t12}
! s5{p0(o0),p0(o2),p1(o0),p1(o2),p1(o3)}: appl={Up(o1,o0)/t13,Right(o2,o3)/t14,Down(o3,o1)/t15,Left(o2,o3)/t16}
! s6{p0(o0),p0(o2),p1(o1),p1(o2),p1(o3)}: appl={Up(o0,o2)/t17,Right(o2,o3)/t18,Down(o1,o0)/t19,Left(o2,o3)/t20}
! s7{p0(o0),p0(o2),p1(o0),p1(o1),p1(o3)}: appl={Right(o2,o3)/t21,Down(o0,o2)/t22,Left(o2,o3)/t23}
! s8{p0(o0),p0(o2),p0(o3),p1(o0),p1(o1),p1(o2)}: appl={Up(o3,o1)/t24,Left(o3,o1)/t25}
! s9{p0(o0),p0(o2),p0(o3),p1(o0),p1(o2),p1(o3)}: appl={Up(o1,o0)/t26,Down(o3,o1)/t27,Left(o3,o1)/t28}
!s10{p0(o0),p0(o2),p0(o3),p1(o1),p1(o2),p1(o3)}: appl={Up(o0,o2)/t29,Down(o1,o0)/t30,Left(o3,o1)/t31}
!s11{p0(o0),p0(o2),p0(o3),p1(o0),p1(o1),p1(o3)}: appl={Down(o0,o2)/t32,Left(o3,o1)/t33}
!---------
\ActionSchema{Up(y,ny):}
 Static: U0(y),B0(y,ny)
 Pre: -NatY(y)
 Eff: NatY(y),-NatY(ny)

\ActionSchema{Right(x,nx):}
 Static: U1(x),U1(nx),B0(x,nx)
 Pre: unaryEncodingX(x),-unaryEncodingX(nx)
 Eff: unaryEncodingX(nx)

\ActionSchema{Down(py,y):}
 Static: U0(py),B0(py,y)
 Pre: NatY(py),-NatY(y)
 Eff: -NatY(py),NatY(y)

\ActionSchema{Left(n,nx):}
 Static: U0(nx),U1(n),B0(n,nx)
 Pre: unaryEncodingX(n),-unaryEncodingX(nx)
 Eff: -unaryEncodingX(n)
\end{Verbatim}

To illustrate the flexibility of the approach, we also show below a first-order representation that is learned from the input graph $G$ that only has 2 labels;
i.e., the labels \atom|Right| and \atom|Left| are  replaced by the label \atom|Horiz|, and the labels \atom|Up| and \atom|Down| by the label \atom|Vert|.

\begin{Verbatim}[label=Grid with 2 labels (ref. 1713)]
!experiments/exp10p-exact-v2/grid1/exp10p_enc4t2_grid1_n1713/theory.cnf 56180 226734 0.320299 1.158141 33932 True experiments/exp10p-exact-v2/grid1/exp10p_enc4t2_grid1_n1713/vtheory_ON_4_grid_v1_4x4.cnf 28982 140004 0.193777 0.142614 33932 True experiments/exp10p-exact-v2/grid1/exp10p_enc4t2_grid1_n1713/vtheory_ON_6_grid_v1_5x6.cnf 129460 548685 0.79535 0.749544 78448 True
!grid1/exp10p_enc4t2_grid1_n1713
\ActionSchema{Horiz(from,to):}
 Static: B2(from,to)
 Pre: atX(from),-atX(to)
 Eff: -atX(from),atX(to)

\ActionSchema{Vert(to,from):}
 Static: B0(to,from)
 Pre: -atY(to),atY(from)
 Eff: atY(to),-atY(from)
\end{Verbatim}

The inferred static predicates \atom|B2| and \atom|B0| capture the horizontal and vertical adjacency relations respectively.

\section{DISCUSSION}

We have shown how  to learn  first-order symbolic representations for  planning
from graphs that only encode the structure of the state space while providing no
information  about the structure of states or actions.
While the formulation of the representation learning problem and
its solution are  very different from those used in deep (reinforcement) learning approaches, there are some
commonalities: we are fitting a parametric representation in the form of theories $T_\alpha(G_{1:n})$ to  data in the form of labeled graphs $G_1, \ldots, G_n$.
The parameters come in two forms: as the  vector of hyperparameters $\alpha$ that bounds  the set of possible first-order planning domains $D$
and the number of objects in each of the instances $P_i = \tup{D,I_i}$, and the boolean variables in the theory  $T_\alpha(G_{1:n})$ that bound
the possible   domains  $D$ and instances $P_i$. The formulation makes room and  exploits a strong structural prior or bias; namely, that
the set of possible domains can be bounded by a small number of hyperparameters with small values (number of action schemas and predicates,
arities, etc).  Lessons  learned, possible extensions, limitations, and challenges are briefly discussed next.

\medskip

\noindent \textbf{Where (meaningful) symbols come from?} We provide a crisp technical answer to this   question in the setting
of planning where meaningful first-order symbolic representations are obtained from non-symbolic inputs in the form of plain state graphs.
In the process,  objects and relations that  are not  given as part of the inputs are learned. The choice of a first-order target language
(lifted STRIPS with negation) was crucial. At the beginning of this work, we tried to learn the (propositional) state variables of a
single instance from the same inputs, but   failed to  obtain the intended variables. Indeed, looking for propositional representations
that minimize the number of variables or the number of variables that change, result in $O(\log|S|)$ variables (where $|S|$ is the number of states)
and  so-called Gray codes, that are not meaningful. The reuse of actions and relations as captured by first-order representations did the trick. 

\medskip

\noindent  \textbf{Traces vs.\ complete graphs.}   The  inputs in our formulation are not  observed traces  but complete graphs.
This distinction, however, is not critical when the graphs required for learning are small.   Using Pearl's terminology \cite{pearl:why},  the input graphs can be regarded
as defining  the complete   space of possible causal interventions that  allow us to recover the   {causal structure} of the  domain  in a first-order language.
The formulation and the SAT encoding, however, can be adjusted in a simple manner  to account for incomplete graphs  where only certain nodes are marked
or assumed to  contain  all of their children. 

\medskip

\noindent \textbf{Non-determinism.}   For learning representations of non-deterministic actions,  the inputs must  be  changed  from OR graphs
   to AND-OR graphs.    Then the action schemas that account for transitions linked by AND nodes must be forced to take the same arguments and the  same preconditions.
   In contrast  to other approaches for learning stochastic action models \cite{kaelbling:stochastic},     this method does not require symbolic inputs but the
   structure of the  space in the form of an AND-OR graph.

    \medskip
    
  \noindent  \textbf{Noise.} The learning approach produces crisp representations from crisp, noise-free inputs.
    However, limited amount of ``noise'' in the form of wrong transitions or labels can  be handled
    at a computational cost by casting the learning task  as an   optimization problem, solvable with  Weighted Max-SAT solvers instead of SAT solvers.

    \medskip

      \noindent  \textbf{Representation learning vs.\ grounding.} The proposed  method learns first-order representations from the structure of the state space,  not from
    the structure of states as displayed for example in  images \cite{asai:latplan,asai:fol}. The latter approaches are less likely
    to generate crisp  representations due to the dependence on images,  but at the same time, they  deal  with two problems at the same time:
    representation learning  and representation (symbol)  grounding  \cite{symbol-grounding}. Our approach deals with  the former problem only;
    the second problem  is for future work.

            \medskip

         \noindent \textbf{Learning or synthesis?} The SAT formulation is used to learn a representation from one or more input graphs corresponding to one or more domain instances.
         The resulting first-order domain representation is  correct for these instances but not  necessarily for other  instances. The more compact the domain representation the more likely
         that it generalizes to other instances, yet studying the conditions under which  this generalization would be correct with high probability
         is beyond the scope of this work.
             
\section{CONCLUSIONS}

We have shown that it is possible to learn first-order symbolic representations for planning  from non-symbolic data
in the form of graphs that only capture the structure of the state space.
Our learning approach is grounded in the simple,  crisp, and powerful principle
of finding a simplest model that is able to explain the structure of the input graphs.
The empirical results show that a number of subtle first-order encodings with static and dynamic predicates
can be obtained in this way.
We are not aware of other approaches that can derive first-order symbolic representations of this type without some
information about the action schemas, relations, or objects.
There are many performance improvements to be pursued in particular regarding to the SAT encoding and the scalability of the approach,
the search in the (bounded) hyperparameter space, and the ranking and  selection of the simplest solutions.
Extensions for dealing  with partial  observations will  be pursued as well.

%

\section*{ACKNOWLEDGEMENTS}

The work was done while B. Bonet  was at the  Universidad Carlos III,  Madrid, on
a sabbatical leave, funded by a Banco Santader--UC3M Chair of Excellence Award.
H. Geffner's work  is partially funded by grant  TIN-2015-67959-P from  MINECO, Spain,
and a grant from the Knut and Alice Wallenberg (KAW) Foundation, Sweden.

  
\bibliography{control}
\bibliographystyle{ecai}

\end{document}